\documentclass[sigconf]{acmart}

\AtBeginDocument{%
  }

\copyrightyear{2025}
\acmYear{2025}
\setcopyright{acmlicensed}\acmConference[CIKM '25]{Proceedings of the 34th ACM International Conference on Information and Knowledge Management}{November 10--14, 2025}{Seoul, Republic of Korea}
\acmBooktitle{Proceedings of the 34th ACM International Conference on Information and Knowledge Management (CIKM '25), November 10--14, 2025, Seoul, Republic of Korea}
\acmDOI{10.1145/3746252.3761320}
\acmISBN{979-8-4007-2040-6/2025/11}

\usepackage{amsmath}
\usepackage{amsthm}
\usepackage{diagbox}
\usepackage{multirow}
\usepackage{enumitem}
\usepackage{bm}
\usepackage[ruled, lined, linesnumbered, commentsnumbered, longend]{algorithm2e}
\usepackage{xcolor}
\usepackage{mathtools}
\usepackage{graphicx}
\usepackage{subfigure}
\usepackage{float}
\usepackage{adjustbox}
\newcommand{\defeq}{\vcentcolon=}
\usepackage{algorithmic}
\DeclareMathOperator{\diag}{diag}

\newcommand{\cc}{\mathbf{c}}
\newcommand{\e}{\mathbf{e}}

\newcommand{\x}{\bm{x}}
\newcommand{\f}{\bm{f}}

\newcommand{\uu}{\bm{u}}
\newcommand{\vvec}{\bm{v}}

\newcommand{\p}{\mathbf{P}}

\newcommand{\I}{\mathbf{I}}
\newcommand{\A}{\mathbf{A}}
\newcommand{\B}{\mathbf{B}}

\newcommand{\W}{\mathbf{W}}
\newcommand{\D}{\mathbf{D}}
\newcommand{\F}{\mathbf{F}}
\newcommand{\LL}{\mathbf{L}}
\newcommand{\UU}{\mathbf{U}}
\newcommand{\CC}{\mathbf{C}}
\newcommand{\VV}{\mathbf{V}}

\newcommand{\Rn}{\mathbb{R}}

\DeclareMathOperator{\tr}{Tr}
\DeclareMathOperator{\Vol}{vol}

\newtheorem{Lemma}{Lemma}

\newcommand{\phu}[1]{\textcolor{black}{#1}}

\begin{document}

\title{FairAD: Computationally Efficient Fair Graph Clustering via Algebraic Distance}

\author{Minh Phu Vuong}
\affiliation{
  \institution{Texas State University}
  \city{San Marcos}
  \state{TX}
  \country{USA}
}
\email{cty13@txstate.edu}

\author{Young-Ju Lee}
\affiliation{
  \institution{Texas State University}
  \city{San Marcos}
  \state{TX}
  \country{USA}
}
\email{yjlee@txstate.edu}

\author{Iv\'an Ojeda-Ruiz}
\affiliation{
  \institution{Lamar University}
  \city{Beaumont}
  \state{TX}
  \country{USA}
}
\email{iojedaruiz@lamar.edu}

\author{Chul-Ho Lee}
\affiliation{
  \institution{Texas State University}
  \city{San Marcos}
  \state{TX}
  \country{USA}
}
\email{chulho.lee@txstate.edu}

\begin{abstract}
Due to the growing concern about unsavory behaviors of machine learning models toward certain demographic groups, the notion of `fairness' has recently drawn much attention from the community, thereby motivating the study of fairness in graph clustering. Fair graph clustering aims to partition the set of nodes in a graph into $k$ disjoint clusters such that the proportion of each protected group within each cluster is consistent with the proportion of that group in the entire dataset. It is, however, computationally challenging to incorporate fairness constraints into existing graph clustering algorithms, particularly for large graphs. To address this problem, we propose FairAD, a computationally efficient fair graph clustering method. It first constructs a new affinity matrix based on the notion of algebraic distance such that fairness constraints are imposed. A graph coarsening process is then performed on this affinity matrix to find representative nodes that correspond to $k$ clusters. Finally, a constrained minimization problem is solved to obtain the solution of fair clustering. Experiment results on the modified stochastic block model and six public datasets show that FairAD can achieve fair clustering while being up to 40 times faster compared to state-of-the-art fair graph clustering algorithms.
\end{abstract}



\maketitle

\section{Introduction}
Recent advancements in machine learning (ML) have enabled its integration into decision-critical applications across various domains, including finance, healthcare, education, and law enforcement. Despite their significant capabilities, ML algorithms are susceptible to biases present in datasets, resulting in potentially \textit{unfair} outcomes for certain demographic groups~\cite{f:2015}. Thus, fairness criteria have been introduced into ML problems, ranging from supervised and unsupervised learning settings~ \cite{x:23, c:17, b:19, k:19, w:23, dong2022edits, du2020fairness, fan2021fair, wang2022unbiased} to semi-supervised and self-supervised settings~\cite{z:20,b:21,z:22}, to eliminate unwanted algorithmic bias and develop fair ML models.

Fairness refers to the unbiased treatment of individuals or groups across various demographic categories, such as race, gender, age, and socioeconomic status. In general, fairness can be incorporated in an ML problem by introducing a fairness regularizer term to its objective function; formulating an optimization problem with explicit fairness constraints; or post-processing the output of a model to account for the fairness. As a result, it leads to an unbiased outcome in the target ML task or the representations that are invariant to protected attributes or have feature distributions that are statistically indistinguishable across demographic groups. 

Fairness has been characterized by several concepts, including individual fairness \cite{jung20, vakilian2022improved}, group fairness \cite{dwork2012fairness}, and counterfactual fairness \cite{kusner2017counterfactual}. Individual fairness requires similar individuals to receive similar outcomes, while counterfactual fairness demands that an individual’s outcome remains unchanged if only their protected attribute were hypothetically modified, with all the other features being held constant. Group fairness ensures a fairly proportional representation across demographic groups. Given the prevalence of demographic groups in real-world datasets, ensuring group fairness has become a critical requirement, especially when it comes to clustering applications.

Chierichetti et al.~\cite{c:17} pioneered the integration of fairness into $k$-center and $k$-median clustering algorithms. They introduced the concept of fairness by ensuring that the proportion of each demographic group within each cluster is consistent with its overall proportion in the dataset. Backurs et al.~\cite{b:19} extended their approach to efficiently handle larger datasets with near-linear running time. More recently, several works have addressed fair $k$-clustering in the presence of outliers \cite{almanza2022k, amagata2024fair}. They first identify a subset of points as outliers to remove and partition the remaining data so that each cluster preserves the overall demographic proportions. These methods, however, have been predominantly applied to clustering tasks in the Euclidean space, but clustering problems also often occur in the context of graph data.

Graph clustering is a fundamental problem and has been extensively studied in the literature. Among others, spectral clustering~\cite{s:00} has been the most popular unsupervised graph-clustering algorithm as it is developed in a principled way to find the optimal solution to a well-defined graph cut problem. It has also been actively extended to variants of the graph cut problem for various reasons, e.g., improving the quality of clustering with assistive input from the user~\cite{y:01,y:04,x:09,n:11,w:14,c:16}. For example, Xu et al. \cite{x:09} incorporate linear constraints to the objective of spectral clustering, while Wang et al. \cite{w:14} enforce prior knowledge via must-link and cannot-link constraints. These constraint‑driven methods are especially effective for image segmentation applications, where a small set of annotated pixels helps the algorithm produce more accurate segmentation boundaries.

Kleindessner et al.~\cite{k:19} introduced a mathematical framework that imposes the notion of fairness as additional linear constraints into the problem of spectral clustering. Their algorithm, which we name as FairSC, however, faces scalability issues for larger graphs due to its high computational cost. Wang et al.~\cite{w:23} recently proposed a scalable fair spectral clustering algorithm named sFairSC by reformulating the problem as a projected eigenvalue problem and effectively improving its scalability. While their algorithms improve the balance performance compared to the standard spectral clustering, their frameworks still rely on solving constrained or projected eigenvalue problems due to the fairness constraints. They generally take much longer than solving unconstrained eigenvalue problems as they require computing the nullspace of a fairness matrix or employing the nullspace projection.

We propose FairAD, a computationally efficient \textbf{fair} graph clustering method via \textbf{A}lgebraic \textbf{D}istance. In FairAD, we first construct a new affinity matrix based on the notion of algebraic distance such that the fairness constraints are imposed. We then employ a recursive graph coarsening process on the affinity matrix to find representative nodes that correspond to a given number of clusters. They eventually lead to a simple constrained minimization problem, which can be solved efficiently. We further optimize the implementation of FairAD through several techniques. 

Our contributions can be summarized as follows: 
\begin{itemize}[itemsep=0pt,leftmargin=1.1em,topsep=1pt]
    \item We introduce a novel framework to integrate fairness constraints into the affinity matrix for graph clustering, when constructed based on the algebraic distance.
    
    \item We demonstrate how graph coarsening can be effectively leveraged to convert the problem into a simpler minimization problem, which can be solved efficiently.    
    
    \item We develop a series of implementation optimizations to further improve the efficiency of FairAD.
    
    \item We evaluate the effectiveness and efficiency of FairAD through extensive experiments on the modified stochastic block model and six real-world datasets. The results show that FairAD not only delivers fair clustering but also runs up to $40\times$ faster than state-of-the-art fair graph clustering algorithms.

\end{itemize}

\section{Preliminaries}

Consider an undirected, weighted graph $G \!=\! (V,E)$, where $V\!=\! \{1,2,\dots,n\}$ is the set of nodes and $E$ is the set of edges. Each edge between nodes $i$ and $j$ is associated with a positive weight $W_{i,j} \!>\! 0$. $W_{i,i} \!=\! 0$ for all $i$, and $W_{i,j} = 0$ if nodes $i$ and $j$ are not neighbors. Let $\W \!=\! (W_{i,j})$ be the $n \times n$ weight matrix, which is also called affinity matrix. Let $d_i$ be the degree of node $i$, which is defined as $d_i := \sum_{j \in V} W_{i,j}$, and let $\D \!\defeq\! \diag(d_1,d_2,\dots,d_n)$ be the degree matrix. For a subset of nodes $A \!\subset\! V$, we define $\Vol(A) \defeq \sum_{i\in A}d_i$ to be a volume of $A$. Also, for two subsets $A, B \!\subset\! V$, we define $W(A, B) \defeq \sum_{i \in A, j \in B}W_{i,j}$. The Laplacian and normalized Laplacian matrices of $G$ are defined as $\LL \defeq \D - \W$ and $\overline{\LL} \defeq \D^{-1/2}(\D - \W)\D^{-1/2}$, respectively. 

\vspace{1mm}
\noindent \textbf{Notations.} For an integer $n \!\geq\! 1$, let $[n] \defeq \{1,2,\ldots, n\}$. Let $\bm{1}$ and $\bm{0}$ denote the $n$-dimensional all-one and all-zero column vectors, respectively. Let $\I_k$ be the $k \times k$ identity matrix. For a node subset $A \!\subset\! V$, let $\overline{A}$ denote its complement $V \setminus A$. 

\subsection{Spectral Clustering}

The basic problem of graph clustering is the minimum cut problem, which is to partition $V$ into $k$ disjoint subsets (clusters), i.e., $V = C_1 \cup C_2 \cup \cdots \cup C_k$, such that the sum of the weights of the edges across different clusters is minimized. That is, it is to find $k$ disjoint subsets to minimize 
\begin{equation*}
    \text{Cut}(C_1, C_2, \ldots, C_k) \defeq \frac{1}{2}\sum_{l=1}^k W(C_l, \overline{C}_l).
\end{equation*}
While this problem can be solved easily, it is widely known that its solution does not lead to satisfactory partitions, and it often separates an individual node from the rest of the graph. Thus, its \emph{properly normalized} versions have been introduced and extensively studied in the literature \cite{n:11, v:07}. Among others, the normalized cut (NCut) problem is the most popular problem, and its corresponding `spectral clustering' algorithm is widely used as an effective graph clustering algorithm~\cite{s:00}. 

Specifically, the NCut problem is to minimize 
\begin{equation}\label{eq:ncut}
    \text{NCut}(C_1, C_2, \ldots, C_k) \defeq \frac{1}{2}\sum_{l=1}^k \frac{W(C_l, \overline{C}_l)}{\Vol(C_l)}.
\end{equation}
Consider $k = 2$. Letting $\bm{h}$ be an indicator vector with entries $h_i = 1$ if $i \in C_1$ and $h_i = -1$ otherwise, the NCut problem in (\ref{eq:ncut}) becomes $\min_{h_i \in \{1 , -1\}} \text{NCut}(\bm{h})$, which can also be written as 

\begin{equation}\label{ncut_k2}
\begin{aligned}
    \min_{u_i \in \{\sigma,-1/\sigma\}} \uu^\top \LL \uu & \\ 
    \text{subject to } \uu^\top\D\uu=\Vol(V) 
    & \text{ and } \uu^\top\D\bm{1} = 0,
\end{aligned}
\end{equation}
where $\sigma$ is some positive constant. Since this problem is NP-hard~\cite{w:93}, by relaxing $\bm{u}$ to take arbitrary real values and substituting $\bm{v} \defeq \D^{1/2}\bm{u}$, we have the following relaxed problem:
\begin{equation}\label{norcut}
\begin{aligned}
    \min_{\vvec \in \Rn^n} \vvec^\top \overline{\LL} \vvec & \\ 
    \text{subject to } \|\vvec\|^2 =\Vol(V) 
    & \text{ and } \vvec^\top\D^{1/2}\bm{1} =0.
\end{aligned}
\end{equation}

This boils down to finding the eigenvector corresponding to the smallest non-zero eigenvalue of the normalized  Laplacian $\overline{\LL}$ and evaluating the ``sign" of each component of the eigenvector to partition the graph into two clusters.

For $k > 2$, by repeating the similar arguments as above, we can relax the $k$-way NCut problem as the following standard trace minimization problem, which is to find the $n \times k$ partition matrix $\mathbf{V}=[\vvec_1, \vvec_2, \dots, \vvec_k]$ to minimize the trace:
\begin{equation}\label{eq:ncut_k}
    \min_{\mathbf{V}\in \Rn^{n\times k}} \tr(\mathbf{V}^\top \overline{\LL} \mathbf{V}) ~~\text{subject to}~ \mathbf{V}^\top\mathbf{V}=\I_k, 
\end{equation}
where $\tr(\mathbf{V}^\top \overline{\LL}\mathbf{V}) = \sum_{i=1}^k \vvec_i^\top \overline{\LL}\vvec_i$. It is also known that the solution to the relaxed $k$-way NCut problem in (\ref{eq:ncut_k}) is to find the eigenvectors that correspond to the $k$ smallest eigenvalues of $\overline{\LL}$~\cite{s:00}.

\begin{figure*}[t!]
    \centering
    \includegraphics[width=0.92\textwidth]{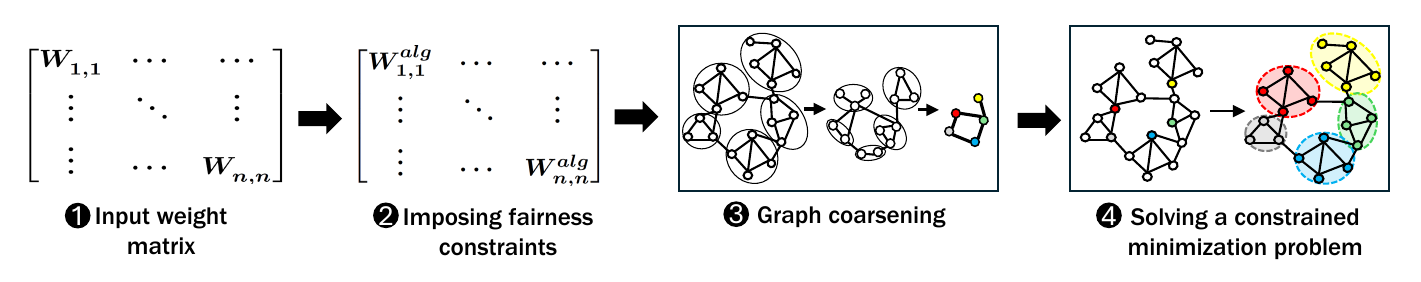}
    \captionsetup{aboveskip=0pt, belowskip=-6pt} 
    \vspace{-2mm}
    \caption{An overview of FairAD.}
    \label{fig:framework}
\end{figure*}
\subsection{Fairness Constraints}

The notion of \emph{fairness} can now be introduced in the context of graph clustering. The goal of fair graph clustering is to ensure that the proportion of nodes in each cluster is identical to the proportion of the population as a whole. Specifically, suppose that nodes are originally divided into $h$ distinct groups, i.e., $V = V_1 \cup V_2 \cup \cdots \cup V_h$. Then, it aims to ensure that, for $s = 1, 2,\ldots, h$ and $l = 1, 2, \ldots, k$,
\begin{equation}\label{eq:bal_cond}
\frac{|V_s \cap C_l|}{|C_l|} = \frac{|V_s|}{|V|}.    
\end{equation}

Let $\f^{(s)} \defeq [f^{(s)}_1, f^{(s)}_2, \ldots, f^{(s)}_n]$ be the group indicator vector for $V_s$, which has elements $f^{(s)}_i = 1$ if $i \in V_s$ and $f^{(s)}_i = 0$ otherwise. Also, let $\F = (F_{i,s})$ be an $n \times (h\!-\!1)$ matrix with elements $F_{i,s} \defeq f^{(s)}_i - |V_s|/|V|$ for $s \in [h-1]$ and $i \in [n]$. Then, as shown in~\cite{k:19,w:23}, a partition $V = C_1\cup C_2\cup \cdots \cup C_k$ is \emph{fair} if and only if its corresponding $n \times k$ partition matrix $\mathbf{V} = [\vvec_1, \vvec_2, \dots, \vvec_k]$ satisfies 
\begin{equation}\label{eq:linear_cons}
\F^\top \mathbf{V}= \bm{0}_{(h\!-\!1) \times k}, 
\end{equation}
where $\bm{0}_{(h-1) \times k}$ is the all-zero matrix of dimension $(h-1) \times k$. In other words, the fairness constraints in (\ref{eq:bal_cond}) are equivalent to the linear constraints in (\ref{eq:linear_cons}). Therefore, the problem of \emph{fair} spectral clustering now becomes 
\begin{equation}\label{eq:fsc_prob}
    \min_{\mathbf{V}^\top\mathbf{V} = \I_k,~ \F^\top \mathbf{V} = \bm{0}_{(h-1) \times k}} \tr (\mathbf{V}^\top \overline{\LL} \mathbf{V}),
\end{equation}
which is imposing the linear constraints in (\ref{eq:linear_cons}) into the problem of spectral clustering in (\ref{eq:ncut_k}). 

To solve this problem efficiently, novel fair spectral clustering algorithms, i.e. FairSC and sFairSC, have been developed \cite{k:19,w:23}. While they improve the balance performance compared to spectral clustering, their frameworks still rely on solving constrained or projected eigenvalue problems due to the fairness constraints. They generally take much longer than solving unconstrained eigenvalue problems as they require computing the nullspace of $\F$ or employing the nullspace projection. As shall be demonstrated through the experiments, their computational time grows quickly with increasing size of the graph. Therefore, there is a need for an efficient and scalable approach for fair graph clustering.

\section{Proposed Method: FairAD}
In this section, we introduce FairAD, a computationally efficient fair graph clustering method. 
We first construct a new affinity matrix based on the notion of algebraic distance, where the fairness constraints are imposed. We then perform a recursive graph coarsening process on this affinity matrix to find representative (or anchor) nodes that correspond to $k$ clusters. We finally determine which cluster each node in the original graph belongs to by solving a relaxed $k$-way graph cut problem where the representative nodes are used as additional linear constraints. In addition to the operations of FairAD, we also explain a set of implementation optimizations made to speed up FairAD in practice. Figure~\ref{fig:framework} illustrates an overview of FairAD.

\subsection{Imposing Fairness Constraints}\label{se:constraints}

For a given graph with the affinity (weight) matrix $\W$, as the first step of FairAD, we construct a new affinity matrix such that the fairness constraints are imposed. To this end, we propose to use the algebraic distance, which was originally developed in~\cite{r:11} to measure the strength of a connection between each node pair in the graph and to construct a new affinity matrix to achieve better solutions to partitioning problems, albeit not under fairness constraints. We below explain how the process of computing the algebraic distance can be modified to impose the fairness constraints into a new affinity matrix.

We begin with the definition of algebraic distance. Let $\x_1, \x_2, \ldots, \x_R$ be the $n$-dimensional \emph{test} vectors. Each test vector $\x_r$ is obtained by running $\tau$ Jacobi relaxation iterations \cite{saad2003iterative} on $\LL \x_r = \bm{0}$, where $\LL= \D-\W$ is the (unnormalized) Laplacian matrix. Starting from a random vector $\x_r^0$, each Jacobi relaxation iteration on $\LL\x_r=\bm{0}$ leads to
\begin{equation*}
    \x^{t}_r = \x_r^{t-1} + \mathbf{D}^{-1}(\bm{0}-\mathbf{L}\x_r^{t-1})= \mathbf{D}^{-1}\W\x_r^{t-1}, \; t=1,2,\dots,\tau-1,
\end{equation*}
and we finally have a test vector $\x_r$ at $t = \tau$. Intuitively, in each iteration, the value of each node is updated based on the weighted average of the values of its neighbors. This iterative process effectively smooths out the values of the nodes that are strongly connected, while preserving the differences across the values of the weakly connected nodes. Then, the algebraic distance between nodes $i$ and $j$ is defined as
\begin{equation}\label{eq:ad}
    s(i,j) =\max_{r=1,2,\ldots,R} |x_{r,i}-x_{r,j}|,
\end{equation}
where $x_{r,i}$ is the $i$-th element of test vector $\x_r$. 
Next, a new affinity matrix, say, $\W_{\text{alg}} \defeq(W^{\text{alg}}_{i,j})$, is constructed based on the algebraic distance in (\ref{eq:ad}) as follows~\cite{r:11}: For all $i,j$, 
\begin{equation}\label{eq:walg}
    W^{\text{alg}}_{i,j} = \exp(-s(i,j)).
\end{equation}

We here aim to incorporate the fairness constraints in (\ref{eq:linear_cons}) into the process of computing the algebraic distance in \eqref{eq:ad}. 
Specifically, our approach is to impose the fairness constraints in (\ref{eq:linear_cons}) at \emph{each} Jacobi relaxation iteration so that its resulting test vector, say $\x^t_r$, at iteration $t$ satisfies the fairness constraints, i.e., $\F^\top\x^t_r=\bm{0}$. For brevity, we drop the subscript $r$ as the test vectors are obtained in the same way. 

First, observe that the vector $\x^{t}$ at the $t$-th Jacobi relaxation iteration is given by 
\begin{equation*}\label{eq:jacobi}
 \x^{t} = \mathbf{D}^{-1}\W \x^{t-1},
\end{equation*}
which leads to
\begin{equation}\label{constraints}
    \mathbf{D}\x^{t}=\mathbf{W}\x^{t-1}.
\end{equation}
We then incorporate the fairness constraints $\F^\top\x^t = \bm{0}$ into (\ref{constraints}), which leads to
\begin{equation} \label{ttt}
    \mathbf{D}\x^{t} = \mathbf{W}\x^{t-1} ~~\text{subject to}~ \F^\top\x^t =\bm{0}.
\end{equation}

Let $\bm{b} \defeq \W \x^{t-1}$. Since we solve this system for each $t$, for ease of exposition, we also drop the superscript $t$. Next, we observe that the system in \eqref{ttt} is equivalent to the following quadratic optimization problem:
\begin{equation}\label{eq:min_cons1}
    \min_{\F^\top\x=\bm{0}} \frac{1}{2} \x^\top\mathbf{D}\x -\bm{b}^\top\x.
\end{equation}
We write its Lagrangian function, which is given by
\begin{equation*}
    \mathcal{L}(\x, \lambda)=\frac{1}{2}\x^\top \D\x - \bm{b}^\top\x +\lambda \F^\top \x,
\end{equation*}
where $\lambda$ is the Lagrange multiplier. The KKT conditions applied to this Lagrangian function yield
\begin{equation}\label{saddle}
\begin{pmatrix} 
\mathbf{D} & \F \\ 
\F^\top &  \bm{0} 
\end{pmatrix} 
\begin{pmatrix} 
\x \\ 
\lambda 
\end{pmatrix} = \begin{pmatrix} 
\bm{b} \\ 
\bm{0}
\end{pmatrix},
\end{equation}
which is a system of linear equations. 

\setlength{\textfloatsep}{0pt}
\begin{algorithm}[t]
     
    \SetKwFunction{isOddNumber}{isOddNumber}
    \SetKwInOut{KwIn}{Input}
    \SetKwInOut{KwOut}{Output}

    \KwIn{$\W $, $\D$, $\F$}
    \KwOut{$\x^{\tau}$.}

    {Initialize $\x^0$}.
    
     \For{$t=1$ \textnormal{\textbf{to}} $\tau$}{
        {$\x^{t} \defeq (\mathbf{D}+\mu \F\F^\top)^{-1}\mathbf{W}\x^{t-1}$.}
    }
    \caption{Constrained Jacobi (cJacobi)} \label{al:jacobi}
\end{algorithm}
We leverage the augmented Lagrangian Uzawa method~\cite{uzawa:00} to solve the indefinite system in \eqref{saddle} as it has a fast rate of convergence, implying that just one iteration provides a good approximate solution. Specifically, given $(\x^\ell, \lambda^\ell)$, a new iterate $(\x^{\ell+1}, \lambda^{\ell+1})$ is obtained by solving the following equations:
\begin{align*}
    (\D+\mu \F\F^\top) \x^{\ell+1} &= \bm{b} - \F \lambda^{\ell}, \\ 
    \lambda^{\ell+1} &= \lambda^0 + \mu \F^\top \x^{\ell},
\end{align*}
where $\mu$ is a penalty parameter. It is known that if $\mu$ is sufficiently large, the iterates converge exponentially fast to the solution of \eqref{saddle}. More formally, the following result is known for the convergence of the Uzawa method from \cite{uzawa:00,o:00}: 
\begin{Lemma}\label{lem:error}
Let $(x^0,\lambda^0)$ be a given initial guess, and for $\ell\ge1$ let
$(x^{\ell},\lambda^{\ell})$ be the iterates produced by the
augmented Lagrangian Uzawa method. Denote by $\gamma_0$ the smallest eigenvalue
of $\F^\top\D^{-1}\F$. Then the following holds:
\begin{align*}
    \|\lambda - \lambda^{\ell}\| \;&\le\;
\Bigl(\tfrac{1}{1 \ + \ \gamma_0 \mu}\Bigr)^{\ell}\,\|\lambda - \lambda^0\|, \\ 
    \|x - x^{\ell}\| \;&\le\;\sqrt{1/\mu}\,\|\lambda - \lambda^{\ell-1}\|
\;\le\;
\sqrt{1/\mu}\,
\Bigl(\tfrac{1}{1\,\ + \ \gamma_0\mu}\Bigr)^{\ell}\,\|\lambda - \lambda^0\|.
\end{align*}
\end{Lemma} 

Lemma \ref{lem:error} implies that the Uzawa method converges exponentially fast for a sufficiently large value of $\mu$. Since the factor $\sqrt{1/\mu}(1+ \gamma_0\mu)^{-1}$ decreases monotonically with $\mu$, even a single iteration with $\mu \gg 1$ can yield a good approximate solution. In other words, by applying the Uzawa method to \eqref{saddle}, for a given $(\x^0, \lambda^0)$, we can obtain $(\x^1, \lambda^1)$ in the first iteration as follows:
\begin{align*}
    (\D+\mu \F\F^\top) \x^1 &= \bm{b} - \F \lambda^0, \\ 
    \lambda^1 &= \lambda^0 + \mu \F^\top \x^1,
\end{align*}
Setting $\lambda^0 = 0$ yields
\begin{equation}\label{yyy}
    (\D+\mu \F\F^\top) \x^1 = \bm{b}.
\end{equation}
By Lemma \ref{lem:error}, we can safely use $\x^1$ in \eqref{yyy} as an approximate solution to \eqref{saddle}. That is, we have the following solution to
\eqref{saddle}:
\begin{equation}\label{approx}
    \x \approx (\D+\mu \F\F^\top)^{-1} \bm{b}.
\end{equation}
Thus, by noting that we have dropped the superscript $t$, and since $\bm{b} = \W \x^{t-1}$, the test vector $\x^t$ at iteration $t$ is now obtained by
\begin{equation}\label{cj}
    \x^{t} = (\D+\mu \F\F^\top)^{-1}\mathbf{W}\x^{t-1}.
\end{equation}

The process of imposing the fairness constraints into every Jacobi relaxation iteration to obtain each test vector $\x_r$ is summarized in Algorithm \ref{al:jacobi}. Once we obtain $R$ test vectors, we compute the algebraic distance $s(i,j)$ for each pair of nodes $i$ and $j$ as in (\ref{eq:ad}) and construct a new affinity matrix $\W_{\text{alg}}$ as in (\ref{eq:walg}).

\subsection{Fair Graph Clustering via Algebraic Distance}

From the new affinity matrix $\mathbf{W}_{\text{alg}}$, which now reflects the fairness constraints, the next step of FairAD is to partition the nodes $V$ into $k$ clusters. To this end, we leverage `graph coarsening' to coarsen the (updated) graph with $\mathbf{W}_{\text{alg}}$ in order to identify a small number of representative corresponding to $k$ clusters. They are then used as anchor nodes to guide the final clustering process. Specifically, we finally solve a constrained minimization problem, which is a relaxed $k$-way graph cut problem with having the representative nodes as additional linear constraints.  

\vspace{1mm}
\noindent \textbf{Graph coarsening.} For graph coarsening, we use a coarsening algorithm introduced in~\cite{sh:00}. It is a recursive algorithm, which starts from the finest level and moves towards increasingly coarser levels. Let $G^{(\ell)} = (V^{(\ell)}, E^{(\ell)})$ be the coarse graph at level $\ell$, and let $\W_\ell = (W_{i,j}^{(\ell)})$ be its corresponding affinity (weight) matrix, where $\ell = 0, 1, \dots, \kappa$. We set $G_0 \defeq G$ and $\W_0 \defeq \W_{\text{alg}}$. That is, the finest graph is the graph $G$ with $\W_{\text{alg}}$. Also, $G^{(\kappa)}$ and $\W_\kappa$ are the coarsest graph and its affinity matrix, respectively. We below explain how $G^{(\ell)}$ and $\W_\ell$ are updated at each level $\ell$.

The coarsening algorithm begins by initializing $V^{(\ell)}$ as a singleton set containing only the first node, say, $n_1\in V^{(\ell-1)}$. That is, $V^{(\ell)} \defeq \{n_1\}$. We then repeatedly check if the next node $n_i\in V^{(\ell-1)}$ is `weakly' connected to the ones that have been added in $V^{(\ell)}$ by evaluating the following inequality:
\begin{equation}
    \max_{j\in V^{(\ell)}} W_{n_i, j}^{(\ell-1)} \le \alpha \sum_{j'\in V^{(\ell-1)}} W_{n_i, j'}^{(\ell-1)},
\end{equation}
where $\alpha$ is the coarsening parameter. The value of $\alpha$ is generally chosen to be much smaller than one, i.e., $\alpha \ll 1$, and our choice of the value shall be explained later in the experiments. If the inequality holds, it implies that node $n_i$ does not have a strong connection with any of the previously added nodes in $V^{(\ell)}$. As a result, node $n_i$ is treated as a `sufficiently independent' node and added to $V^{(\ell)}$, i.e., $V^{(\ell)} \defeq V^{(\ell)}\cup\{n_i\}$. Otherwise, $V^{(\ell)}$ remains unchanged. This coarsening process is repeated until all the nodes $V^{(\ell-1)}$ are evaluated. Note that the intuition behind this process is to eliminate \emph{mutually strongly connected} nodes at each level.
\setlength{\textfloatsep}{0pt}
\begin{algorithm}[t]
     
    \SetKwFunction{isOddNumber}{isOddNumber}
    \SetKwInOut{KwIn}{Input}
    \SetKwInOut{KwOut}{Output}

    \KwIn{$\mathbf{W}_{\text{alg}}$, \# of coarse levels $\kappa$}

    \KwOut{Coarse graphs $\{G_{\ell}\}_{\ell=1}^\kappa$ with $\{\W_{\ell}\}_{\ell=1}^\kappa$}

    {$\W_0\defeq \W_{\text{alg}}$, $V^{(0)}\defeq V$.}
    
    \For{each coarse level $\ell = 1,2,\ldots,\kappa$}{
         {$\eta \defeq |V^{(\ell-1)}|$.}
         
        {$V^{(\ell)}\defeq \{ n_1\}$}.
        
        \For{$i = 2,3, \ldots, \eta$}{
            \vspace{1mm}
            \If{$\max_{j\in V^{(\ell)}}W_{n_i,j}^{(\ell-1)}\leq\alpha\sum_{j'\in V^{(\ell-1)}}W_{n_i,j'}^{(\ell-1)}$}{$V^{(\ell)}\defeq V^{(\ell)}\cup \{n_i\}$.}
            
        }
        {Compute $\p_{\ell}$ by (\ref{eq:p1}).}
        
        {$\W_{\ell} \defeq \p_{\ell}^\top \W_{\ell-1} \p_{\ell}$.}
    }
    \caption{Coarsening} \label{al:co}
\end{algorithm}

Once the coarsening process is complete at level $\ell$, we construct a $|V^{(\ell-1)}| \times |V^{(\ell)}|$ interpolation matrix $\p_{\ell} \defeq (P^\ell_{i,j})$, with elements $P^\ell_{i,j}$ given by  
\begin{equation}\label{eq:p1}
    P^\ell_{i,j} \defeq 
   \begin{cases} 
        \frac{W_{i,j}^{(\ell-1)}}{\sum_{j'\in V^{(\ell)}} W_{i, j'}^{(\ell-1)}},   &\text{for } i \in V^{(\ell-1)}, j\in V^{(\ell)}, \\
        1, &\text{for } i\in V^{(\ell)}, i=j, \\
        0, &\text{otherwise.}
    \end{cases}
\end{equation}
This interpolation matrix is then used to obtain the affinity matrix $\W_\ell$ at level $\ell$ as follows:
\begin{equation*}
    \W_{\ell} = \p^T_{\ell} \W_{\ell-1}\p_{\ell}.
\end{equation*}
The entire graph coarsening process is repeated recursively, level by level. It is summarized in Algorithm \ref{al:co}.

\setlength{\textfloatsep}{0pt}
\begin{algorithm}[t]
     
    \SetKwFunction{isOddNumber}{isOddNumber}
    \SetKwInOut{KwIn}{Input}
    \SetKwInOut{KwOut}{Output}

    \KwIn{$\mathbf{W}$,  $\mathbf{D}$,  $\F$, $m$}
    \KwOut{$\vvec_1,\vvec_2,\ldots, \vvec_k$}
    \For{$r = 1$ \KwTo $R$}{
       { $\x_r \leftarrow$ \text{cJacobi}($\W,\D, \F$).}
    }
    { Compute $\mathbf{W}_{\text{alg}} $ using (\ref{eq:walg}) with $\{\x_r\}_{r=1}^R$.}
    
    { $\{G_{\ell}\}_{\ell=0}^\kappa \leftarrow\text{ Coarsening}( \mathbf{W}_{\text{alg}} )$.} 
    
    \For{ each coarse level $\ell = \kappa,\kappa-1,\ldots,0 $}{
        \If{$ |V^{(\ell)}| \geq m $}{

            $\B,\cc \leftarrow \text{SpectralClustering}(G_{\ell}, k)$. \\
            break;
         }
    }
     {${\overline{\LL}}_{\text{alg}} \defeq \D_{\text{alg}}^{-1/2} (\D_{\text{alg}}-\W_{\text{alg}})\D_{\text{alg}}^{-1/2}$.}
     
     {$\A_{\text{alg}} \defeq \overline{\LL}_{\text{alg}}+\mu \B^\top \B$.}
     
     \For {each cluster $i = 1,2,\dots, k$}{
        $\vvec_i \defeq \mu \A_{\text{alg}}^{-1} \B^\top \cc_i.$ 
     } \label{al:res}
    \caption{FairAD} \label{al:al}
\end{algorithm}

\noindent \textbf{Constrained minimization.} The primary objective of graph coarsening is to identify $k$ representative nodes corresponding to $k$ distinct clusters. Intuitively, they are \emph{most weakly} connected to each other, so they could serve as anchor nodes (i.e., each node represents each separate cluster) for the final clustering process. However, a drawback of the graph coarsening algorithm is that we cannot control the exact number of nodes generated at the coarsest level ($\ell = \kappa$). Thus, we instead identify at least $m > k$ representative nodes that correspond to $k$ clusters, where multiple nodes can correspond to the same cluster. Once we obtain coarse graphs $\{G_\ell\}_{\ell=1}^\kappa$ from the coarsening algorithm, we find the smallest coarse graph containing at least $m$ nodes. Note that for a given value of $k$, we set the value of $m$ to be a bit greater than the value of $k$. For example, we use the value of $m$ between 15 and 50 for $k < 10$. 

Specifically, we move towards the finest graph, starting from the coarsest one, and find the first coarse graph containing at least $m$ nodes (Lines 6-7 of Algorithm~\ref{al:al}). We then apply spectral clustering to this coarse graph to obtain $k$ groups of representative nodes (Line 8 of Algorithm~\ref{al:al}). In other words, we first compute the first $k$ eigenvectors of the $m\times m$ normalized Laplacian matrix of the coarse graph, which form a $m\times k$ matrix, and then run $k$-means on its rows to produce $k$ groups of representative nodes. Note that since the coarse graph (with $m$ nodes) is significantly smaller than the original graph (with $n$ nodes), it is not a computational burden to use the spectral clustering at this stage.

Let $m^*$ be the number of identified representative nodes. Our next step is to find the clustering solution such that $m^*$ representative nodes belong to their corresponding clusters. In other words, we impose the representative nodes with their corresponding groups as linear constraints to a relaxed graph cut problem. This problem is given by, for $i = 1,2, \ldots, k$,
\begin{equation}\label{eq:min_cons}
    \min_{\B\vvec_i=\cc_i} \frac{1}{2} \vvec^\top_i \overline{\mathbf{L}}_{\text{alg}} \vvec_i,
\end{equation}
where $\overline{\LL}_{\text{alg}}$ is the normalized Laplacian corresponding to $\mathbf{W}_{\text{alg}}$. From $m^*$ representative nodes, say, $r_1, r_2, \ldots, r_{m^*}$, we define $\B$ and $\cc_i$ to be 
\begin{equation}
    \B = \begin{pmatrix} 
    \e_{r_1}^\top \\ 
    \e_{r_2}^\top  \\
    \vdots \\
    \e_{r_{m^*}}^\top
    \end{pmatrix} \quad \text{and} \quad \cc_i = \begin{pmatrix} 
    \delta_{r_1,i} \\ 
    \delta_{r_2,i}  \\
    \vdots \\
    \delta_{r_{m^*},i}
    \end{pmatrix},
\end{equation}
respectively, where $\e_{i}$ is an $n$-dimensional vector with the $i$-th entry being one and the others being zero, and $\delta_{r,i} = 1$ if node $r$ belongs to cluster $i$ and 0 otherwise. That is, each row of $\B$ is a one-hot vector that represents the location of its corresponding representative node, and $\cc_i$ is a vector to indicate which representative nodes belong to cluster $i$.

We can employ the same technique as used in solving (\ref{eq:min_cons1}) to solve (\ref{eq:min_cons}). First, we write its equivalent indefinite system as
\begin{equation}\label{xxx}
\begin{pmatrix} 
\overline{\mathbf{L}}_{\text{alg}} & \B^\top \\ 
\B &  \bm{0} 
\end{pmatrix} 
\begin{pmatrix} 
\vvec_i \\ 
\lambda 
\end{pmatrix} = \begin{pmatrix} 
\bm{0} \\ 
\cc_i 
\end{pmatrix}. 
\end{equation}
Then, as was done in Section~\ref{se:constraints}, we leverage the Uzawa method to solve the system in \eqref{xxx}. For a given  $(\vvec_i^0, \lambda^0)$, we have $(\vvec_i^1, \lambda^1)$ in the first iteration as follows:
\begin{align*}
    (\overline{\mathbf{L}}_{\text{alg}}+\mu \B^\top\B) \vvec_i^1 + \B \lambda^0 &= \mu \B^\top \cc_i \\ 
    \lambda^1 &= \lambda^0 + \mu (\B\vvec^1_i-\cc_i).
\end{align*}
Setting $\lambda^0=0$ yields
\begin{equation*}\label{yyy2}
    (\overline{\mathbf{L}}_{\text{alg}}+\mu \B^\top \B) \vvec_i^1 = \mu \B^\top\cc_i.
\end{equation*}
By Lemma \ref{lem:error}, we obtain the following approximate solution to \eqref{xxx}:
\begin{equation} \label{uzawa}
    \vvec_i \approx \mu \A_{\text{alg}}^{-1} \B^\top \cc_i,
\end{equation}
where $\A_{\text{alg}} := \overline{\mathbf{L}}_{\text{alg}} + \mu \B^\top \B$. Finally, once we have the solutions $\vvec_1,\vvec_2,\ldots,\vvec_k$, the cluster label of node $j$ is determined by identifying which $\vvec_i$ has the maximum value in its $j$-th entry, i.e., $\arg\max_{i} v_{i,j}$ for $j \in [n]$.

\begin{figure}[t!]
    \centering
    \includegraphics[width=0.47\textwidth]{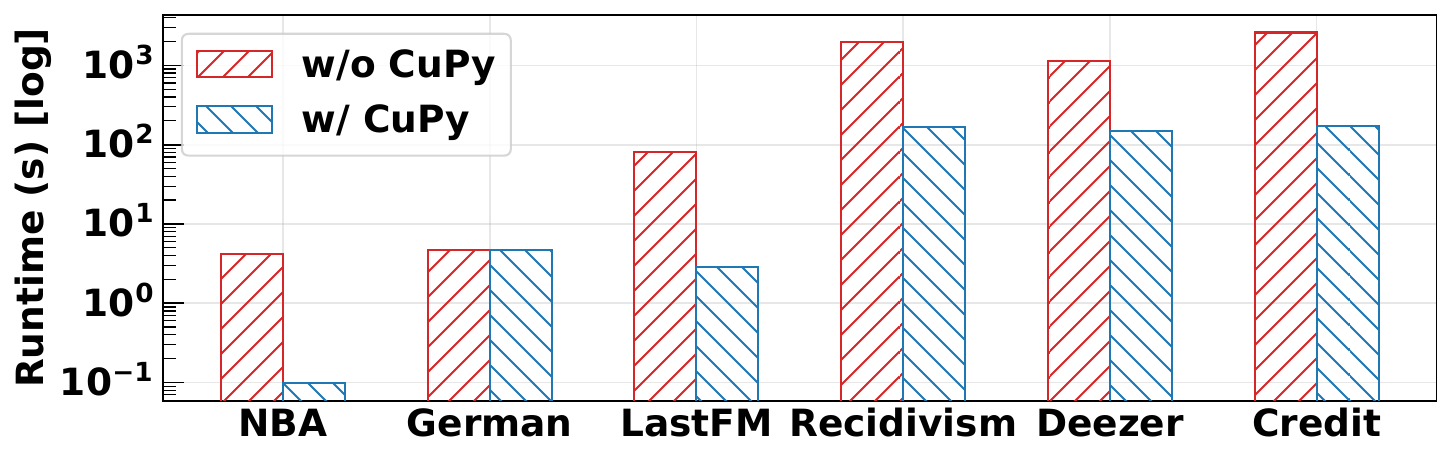}
    \captionsetup{aboveskip=2pt, belowskip=-3pt} 
    \caption{Running time of FairAD with and without CuPy.}
    \label{fig:cupy}
\end{figure}

\begin{figure}[t!]
    \centering
    \includegraphics[width=0.47\textwidth]{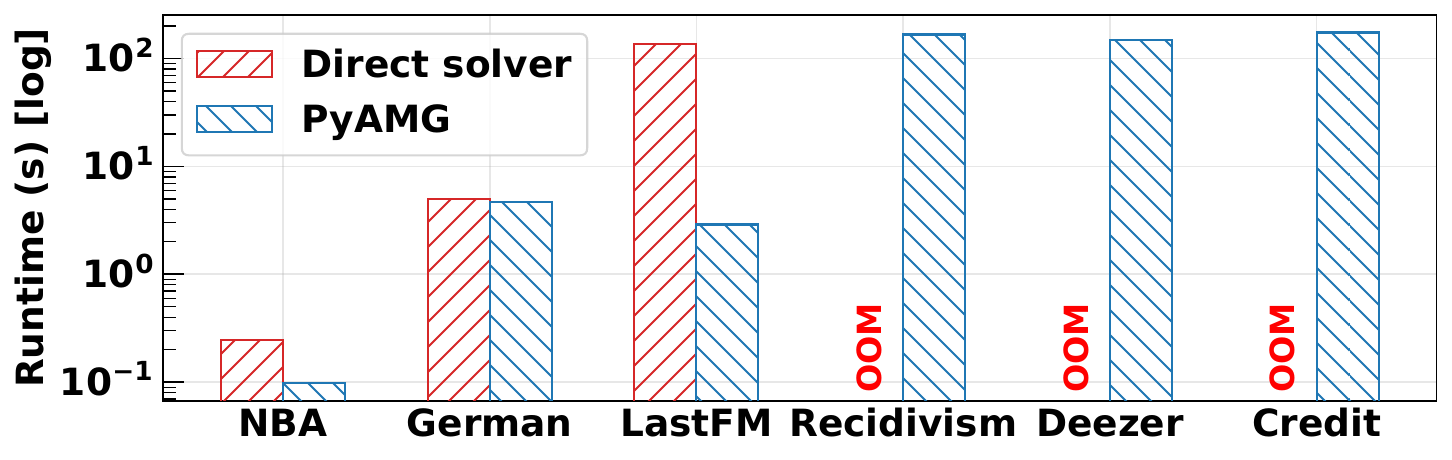}
    \captionsetup{aboveskip=1pt, belowskip=--2pt} 
    \caption{Running time of FairAD with and without PyAMG.}
    \label{fig:pyamg}
\end{figure}

\subsection{Optimized Implementation}

We here explain a set of implementation optimizations used for the implementation of FairAD, which is summarized in Algorithm \ref{al:al}. First, recall that obtaining the test vectors $\x_1, \x_2, \ldots, \x_R$ to construct a new affinity matrix $\W_{\text{alg}}$ requires computing \eqref{cj} iteratively (or running Algorithm~\ref{al:jacobi}), which involves inverting the matrix $(\mathbf{D}+\mu \F\F^\top)$. To efficiently compute $(\mathbf{D}+\mu \F\F^\top)^{-1}$, we employ the Woodbury matrix identity \cite{w:195}. For  given matrices $\A,\CC,\UU$, and $\mathbf{V}$ with the shapes $n\times n, k\times k, n\times k$ and $k\times n$, respectively, the Woodbury matrix identity to compute $(\A+\UU \CC \mathbf{V})^{-1}$ is given by
\begin{equation*}\label{eq:wood}
    (\A+\UU \CC \mathbf{V})^{-1} = \A^{-1}-\A^{-1}\UU(\CC^{-1}+\mathbf{V} \A^{-1}\UU)^{-1}\mathbf{V}  \A^{-1}.
\end{equation*}
Using this identity, we can efficiently compute $(\D +\mu \F \F^\top)^{-1}$ as follows:
\begin{equation*}
    (\D+\mu\F\F^\top)^{-1}=\D^{-1}-\D^{-1}\F(\I_\mu^{-1}+\F^\top\D^{-1}\F)^{-1}\F^\top\D^{-1},
\end{equation*}
where $\I_{\mu} \defeq \mu \I$. Here, since $\D$ is a diagonal matrix, it is straightforward to compute $\D^{-1}$. In addition, $\F$ is a tall matrix, meaning that the inversion $(\I_\mu^{-1}+ \F^\top\D^{-1}\F)^{-1}$ is also easier to compute. This is because it only involves inverting a $k\times k$ matrix, where $k$ is typically much smaller than $n$. That is, the inversion costs $\mathcal{O}(k^3)$ instead of $\mathcal{O}(n^3)$.

Second, we optimize the implementation of FairAD by leveraging CuPy, which is an open source library for GPU-accelerated computing with Python.\footnote{\url{https://docs.cupy.dev/en/v13.2.0/index.html}} In particular, the matrix operations are done much more efficiently since CuPy allows us to fully exploit GPU parallel processing. As a result, the computational efficiency of FairAD can be greatly enhanced. In Figure \ref{fig:cupy}, we compare the running times of our GPU-accelerated implementation of FairAD with CuPy and its CPU-based counterpart. The GPU-accelerated implementation achieves at least an order-of-magnitude speed-up on five of the six datasets. Specifically, we observe a nearly two orders-of-magnitude improvement on NBA and LastFM graphs and roughly an order-of-magnitude improvement on the larger graphs such as Recidivism, Deezer, and Credit.

Third, we optimize the efficiency of the graph coarsening algorithm, which is an integral part of FairAD, by judiciously choosing the order of nodes in $V^{(\ell-1)}$ to be evaluated, as in Line 5 of Algorithm~\ref{al:co}. Instead of simply evaluating the nodes in increasing order of their node IDs, we prioritize the nodes that are strongly connected with others. To this end, we maintain a set of `volumes', denoted by $\bm{\nu}$. It is initially $\bm{\nu} \defeq \bm{1}$ and updated by $\bm{\nu} \defeq \bm{\nu}\p_{\ell}$. The nodes in $V^{(\ell-1)}$ are then evaluated in descending order of their volumes. This way, we can first evaluate more important nodes from a network connectivity perspective at each level.

Finally, to identify the cluster membership of each node in the end, i.e., which cluster each node belongs to, we need to compute the solution $\vvec_i$ as in (\ref{uzawa}), which is equivalent to solving the following linear system: For $i = 1, 2, \ldots, k,$
\begin{equation*}\label{eq:cons}
    \A_{\text{alg}} \vvec_i = \mu \B^\top \cc_i.
\end{equation*}
Note that unlike computing $(\mathbf{D}+\mu \F\F^\top)^{-1}$, it is impractical to directly compute $\A_{\text{alg}}^{-1}$, since $\A_{\text{alg}}$ does not possess the nice property of $(\mathbf{D}+\mu \F\F^\top)$ that allows us to leverage the Woodbury matrix identity. One could solve the linear system using the standard solvers from NumPy or SciPy's linear algebra packages, but they become inefficient for large graphs. Thus, we utilize PyAMG's classical AMG solver~\cite{pyamg23}, which is well-suited for large, sparse systems.\footnote{\url{https://pyamg.readthedocs.io/en/latest/}} In Figure \ref{fig:pyamg}, we compare the running times of the implementation of FairAD when using the PyAMG solver and a direct SciPy solver. For the small graphs such as NBA and LastFM, the implementation of FairAD with PyAMG outperforms the one with the direct solver by up to an order of magnitude. For the larger graphs such as Recidivism, Deezer, and Credit, the implementation with PyAMG completes in roughly 150 seconds, while the one with the direct solver fails with out-of-memory errors (``OOM''). Thus, we can see that the implementation of FairAD based on the PyAMG solver is more computationally efficient and scalable to large graphs.

\section{Experiments}
\phu{\begin{table}[t!]
    \centering
    \caption{Statistics of datasets used in the experiments} \label{tab:ds}
    \vspace{-2mm}
    \begin{adjustbox}{width=0.9\columnwidth,center}
    \begin{tabular}{|c|c|c|c|c|}
        \hline
        \textbf{Dataset} & $\mathbf{|V|}$ & $\mathbf{|E|}$ & \textbf{Sensitive Attribute} & $\mathbf{h}$ \\
        \hline
        NBA & 403 & 10,621 & Country & 2 \\
        German & 1,000 & 21,742 & Gender & 2 \\
        LastFM & 7,624 & 27,806 & Country & 4 \\
        Recidivism & 18,876 &  311,870 & Race & 2 \\
        Deezer & 28,281 & 92,752 & Gender & 2 \\
        Credit & 29,460 & 136,196 & Education & 3 \\
        \hline
    \end{tabular}
    \end{adjustbox}
\end{table}}

\begin{figure*}[!t]
  \setlength{\subfigcapskip}{-4pt}
  \centering
  \subfigure{%
    \includegraphics[width=0.4\textwidth]{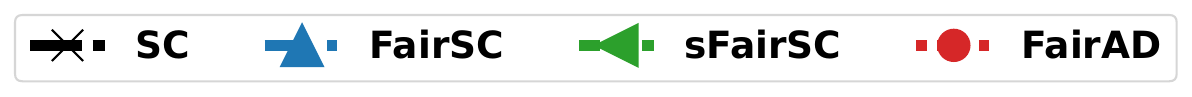}%
 \addtocounter{subfigure}{-1}
  }\\[-4mm]  

  \subfigure[$h=2,k=4$]{%
    \includegraphics[width=0.32\textwidth]{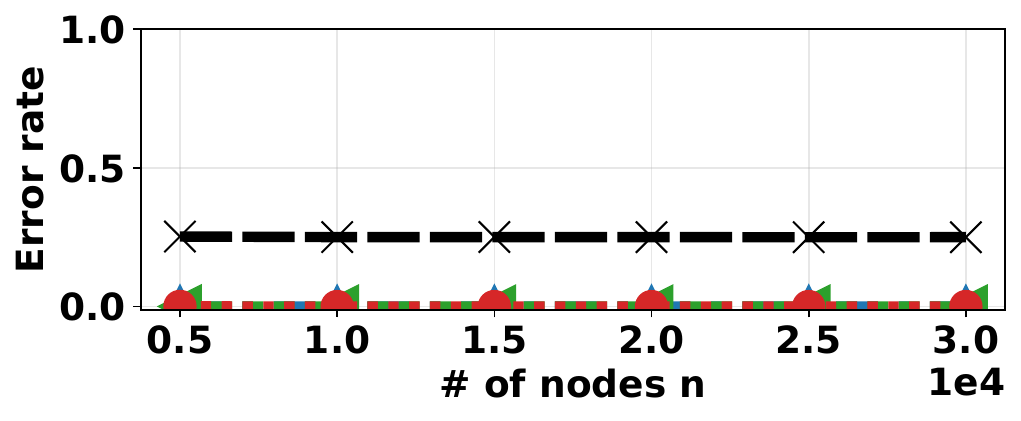}%

  }\hfill
  \subfigure[$h=5,k=5$]{%
    \includegraphics[width=0.32\textwidth]{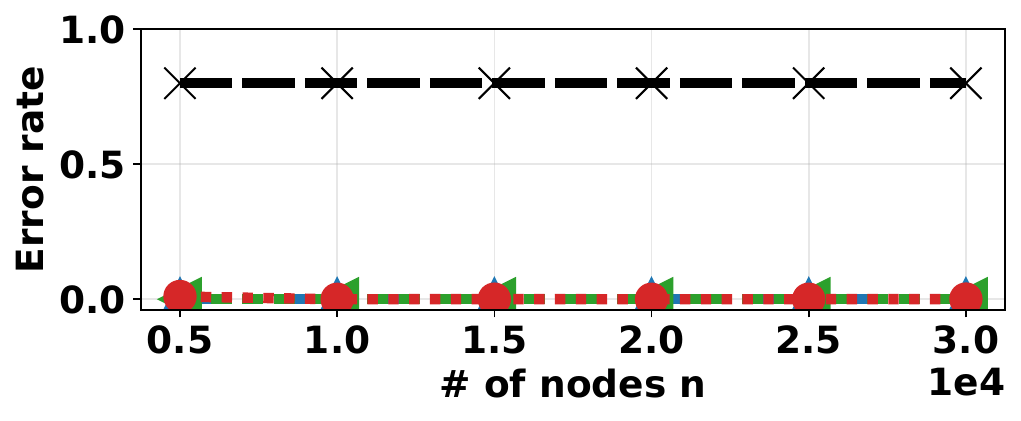}%

  }\hfill
  \subfigure[$h=10,k=5$]{%
    \includegraphics[width=0.32\textwidth]{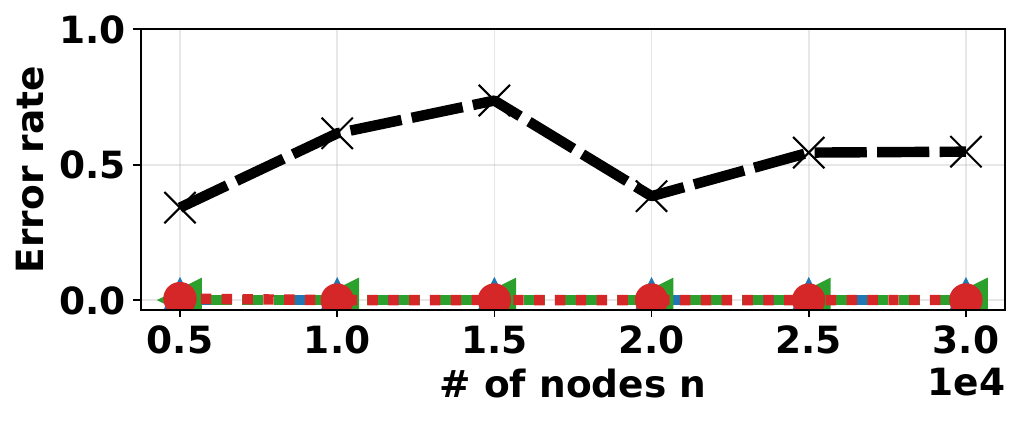}%

  }
  \\
  \vspace{-3mm}
    \subfigure[$h=2,k=4$]{%
    \includegraphics[width=0.32\textwidth]{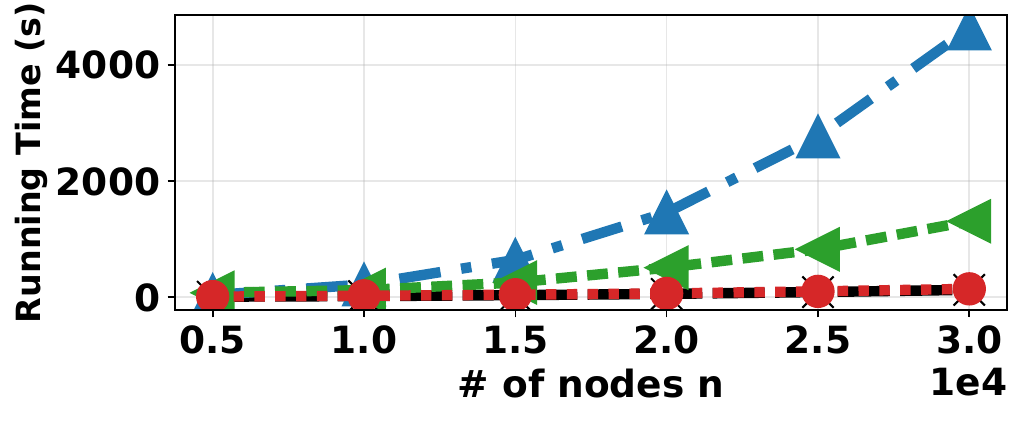}%

  }\hfill
  \subfigure[$h=5,k=5$]{%
    \includegraphics[width=0.32\textwidth]{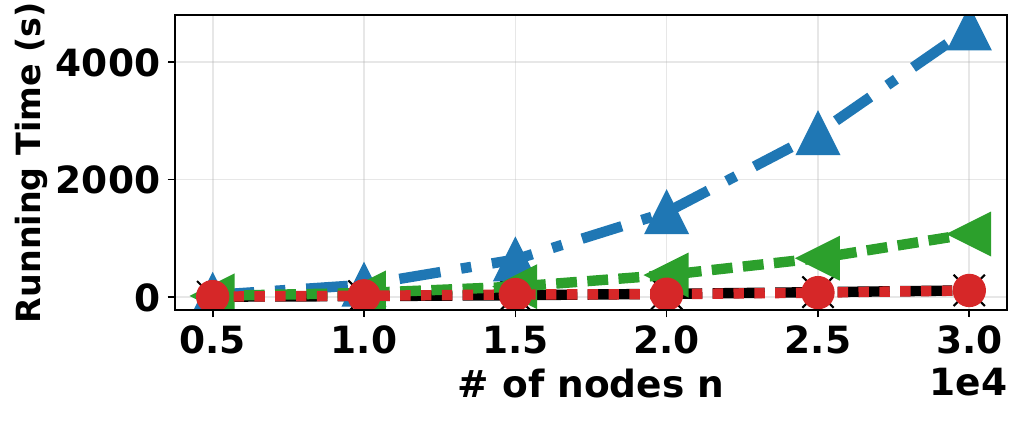}%

  }\hfill
  \subfigure[$h=10,k=5$]{%
    \includegraphics[width=0.32\textwidth]{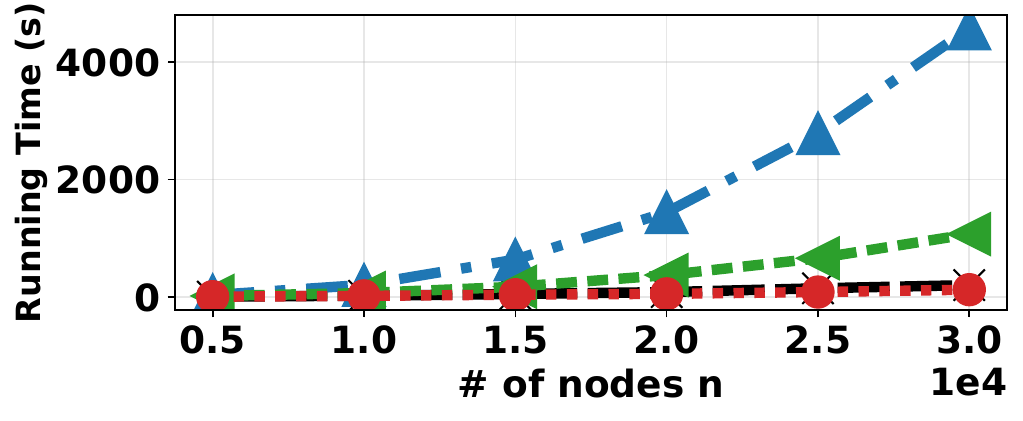}%

  }
  \vspace{-5mm}
  \caption{%
    Error rate in (\ref{er}), shown in the top row, and running time (in seconds), shown in the bottom row, under synthetic networks generated by mSBM with varying values of $h$ and $k$.
  }
  \label{fig:sbm1}
\end{figure*}
\vspace{-1.5ex}
In this section, we provide extensive experiment results to demonstrate the superior performance of FairAD to baseline algorithms, which are state-of-the-art fair clustering algorithms such as FairSC and sFairSC as well as the plain spectral clustering (SC).

\vspace{1mm}
\noindent \textbf{Datasets.} We consider both synthetic and public real-world datasets for performance evaluation. The synthetic dataset is generated based on a modified stochastic block model (mSBM) \cite{k:19}, which has been widely used to generate synthetic networks for clustering and community detection. Suppose that the set of nodes $V$ consists of $h$ groups, i.e., $V = V_1 \cup V_2 \cup \cdots \cup V_h$, and is also partitioned into $k$ ground-truth clusters, i.e., $V  =  C_1 \cup C_2 \cup \cdots \cup C_k$. The synthetic dataset is generated such that the `fairness' condition is satisfied, meaning that the almost same proportion of nodes from each group $V_s$ appears in each cluster $C_l$. In other words, we ensure the proportion $\eta_s \defeq |V_s \cap C_l| / |C_l|$ to be more or less the same for all $s \in [h]$, i.e., $\eta_1\approx\eta_2\approx\ldots\approx\eta_h$, for each $l \in [k]$.

For mSBM, the probability of having an edge between two nodes depends on their membership in the groups and clusters. Thus, we define the probability of having an edge between nodes $i$ and $j$, say, $Q_{i,j}$, to be given by
\begin{equation*}
    Q_{i,j} \defeq 
   \begin{cases}
        a,~ \text{if $i$ and $j$ are in the same group and the same cluster,} \\
        b,~ \text{if $i$ and $j$ are in the same group and different clusters,} \\
        c,~ \text{if $i$ and $j$ are in different groups and the same cluster,} \\
        d,~ \text{if $i$ and $j$ are in different groups and different clusters,} \\
    \end{cases}
\end{equation*}
with $a > b > c > d$, which is to reflect stronger connections within groups and clusters~\cite{k:19,w:23}. Then, the affinity matrix of mSBM is obtained by
\begin{equation*}
    W_{i,j}= \begin{cases}
        \text{Bernoulli}(Q_{i,j}), &\text{if } i\neq j \\
        0, &\text{otherwise,}
    \end{cases}
\end{equation*}
where $\text{Bernoulli}(Q_{i,j})$ is a Bernoulli random variable with probability $Q_{i,j}$.

\begin{table*}[htbp]
  \centering
 \caption{Running times (in seconds) of SC, FairSC, sFairSC, and FairAD on mSBM with $n = 20000$}\label{tab:performance}
 \vspace{-2mm}
  \resizebox{\textwidth}{!}{%
    \begin{tabular}{|l|c|c|c|c|c||c|c|c|c|c||c|c|c|c|c||c|c|c|c|c|}
      \hline
      & \multicolumn{5}{c||}{$k=3$} & \multicolumn{5}{c||}{$k=4$} & \multicolumn{5}{c||}{$k=5$} & \multicolumn{5}{c|}{$k=6$} \\
      \hline
      & $h=2$ & $h=4$ & $h=6$ & $h=8$ & $h=10$ & $h=2$ & $h=4$ & $h=6$ & $h=8$ & $h=10$ & $h=2$ & $h=4$ & $h=6$ & $h=8$ & $h=10$ & $h=2$ & $h=4$ & $h=6$ & $h=8$ & $h=10$ \\
      \hline\hline
      SC  & 44 & 34 & 55 & {53} & 53 & 32 & 22 & 26 & {46} & 47 & 53 & 35 & 46 & 55 & 58 & 79 & 58 & 44 & 39 & 53 \\
      \hline
      FairSC & 1337 & {1215} & 1249 & 1329 & 1357 & 1304 & {1325} & 1361 & 1213 & 1196 & 1137 & 1253 & 1143 & 1156 & 1243 & 1248 & 1313 & 1337 & 1189 & 1179 \\
      \hline
      sFairSC & {546} & 512 & {535} & 636 & 457 & {444} & 499 & {466} & 450 & 526 & 578 & 534 & 698 & 437 & 567 & 642 & 603 & 654 & 433 & 603 \\
      \hline
      FairAD & {57} & 52 & {51} & 52 & 52 & {61} & 60 & {56} & 58 & 55 & 65 & 63 & 64 & 62 & 63 & 81 & 66 & 69 & 63 & 71 \\
      \hline
    \end{tabular}%
    
  }

\end{table*}

\begin{table*}[htbp]
  \centering
      \caption{Running times (in seconds) of SC, FairSC, sFairSC, and FairAD on mSBM with $n = 30000$}
    \label{tab:performance2}
    \vspace{-2mm}
  \resizebox{\textwidth}{!}{%
    \begin{tabular}{|l|c|c|c|c|c||c|c|c|c|c||c|c|c|c|c||c|c|c|c|c|}
      \hline
      & \multicolumn{5}{c||}{$k=3$} & \multicolumn{5}{c||}{$k=4$} & \multicolumn{5}{c||}{$k=5$} & \multicolumn{5}{c|}{$k=6$} \\
      \hline
      & $h=2$ & $h=4$ & $h=6$ & $h=8$ & $h=10$ & $h=2$ & $h=4$ & $h=6$ & $h=8$ & $h=10$ & $h=2$ & $h=4$ & $h=6$ & $h=8$ & $h=10$ & $h=2$ & $h=4$ & $h=6$ & $h=8$ & $h=10$ \\
      \hline\hline

      SC  & 73 & 78 & 86 & {111} & 97 & 111 & 72 & 84 & {111} & 87 & 89 & 89 & 78 &107 & 122 & 82 & 82 & 94 & 107 & 122 \\
      \hline

      FairSC & 4502 & {4287} & 4383 & 4285 & 4472 & 4574 & {4523} & 4567 & 4502 & 4421 & 4439 & 4553 & 4494 & 4529 & 4519 & 4510 & 4332 & 4453 & 4339 & 4379 \\
      \hline

      sFairSC & {1008} & 1036 & {1003} & 963 & 1001 & {1032} & 1006 & {1007}& 962& 1044 & 1099 & 1157 & 1063 & 1076 & 981 & 890 & 862 & 804 & 883 & 789 \\
      \hline

      FairAD & {122} & 116 & {114} & 112 & 114 & {149} & 129 & {133} & 132 & 140 & 140 & 149 & 134 & 130 & 131 & 161 & 149 & 146 & 148 & 158 \\
      \hline
    \end{tabular}%
  }
\end{table*}

In addition, we consider six real-world datasets, namely, NBA \cite{dai2021say}, German \cite{asuncion2007uci}, Recidivism \cite{jordan2015effect}, LastFM \cite{snap:20}, Deezer \cite{snap:20}, and Credit \cite{yeh2009comparisons}, which are all from social networks and contain sensitive attributes. The statistics of the datasets are provided in Table \ref{tab:ds}. We use the largest connected component of each graph.

\vspace{1mm}
\noindent \textbf{Parameter settings.} In our experiments, we set the parameters of mSBM as follows: $a  =  10(\log n/n)^{2/3}$, $b  =  7(\log n/n)^{2/3}$, $c  =  4(\log n/n)^{2/3}$, and $d  = (\log n/n)^{2/3}$. For FairAD, we set $\mu  =  10^9$ and $\alpha  =  10^{-4}$. Both the number of test vectors, $R$, and the number of Jacobi iterations, $\tau$, are set to 10. We observed that the test vectors $\x_1, \x_2, \ldots, \x_R$ often become indistinguishable due to floating point precision limitations. To address this issue, we introduce a scaling parameter $\beta$ to the definition of $W^{\text{alg}}_{i,j}$ in \eqref{eq:walg}, which is now given by $W^{\text{alg}}_{i,j} = \exp(-\beta s(i,j))$, where the value of $\beta$ is chosen to be sufficiently large. This way, we ensure that the difference between the scaled algebraic distances of different node pairs becomes much bigger, thereby leading to informative test vectors. Specifically, we set $\beta=n/\log(n)$ in the experiments.

\subsection{Experiment Setup}
\noindent \textbf{Evaluation metrics.} We use the error rate introduced in \cite{k:19} to quantify the performance of FairAD and baselines on the synthetic networks generated by mSBM. Let $\VV=(V_{i,j})$ be the indicator matrix of the predicted clustering labels, where $V_{i,j}=1$, if node $i$ belongs to cluster $j$, and $0$ otherwise. Similarly, let $\VV^*$ be the indicator matrix for the ground-truth labels generated by the mSBM. Note that the numeric labels $1,2,\dots,k$ are arbitrary, so the same partitions may differ by a permutation of columns. In other words, there exists a permutation matrix $\mathbf{U}$ such that $\VV\mathbf{U}=\VV^*$. To account for this nature, letting $\mathbf{\Pi}_k$ be the set of all $k\times k$ permutation matrices, the error rate is defined as the smallest difference between the permuted prediction $\VV\mathbf{U}$ and the ground truth $\VV^*$, which is given by
\begin{equation}\label{er}
    E(\VV-\VV^*)=\frac{1}{k}\min_{\mathbf{U}\in \mathbf{\Pi}_k}\|{\VV\mathbf{U}-\VV^*}\|.
\end{equation}

For real-world datasets, we use the average balance introduced in~\cite{c:17} as the performance metric. Specifically, for a given group partition $V  =  V_1 \cup \cdots \cup V_h$, having a partition of clusters $V  =  C_1 \cup  \cdots \cup C_k$, we define the balance of cluster $C_l$ as follows: For $l=1,2,\dots,k$, 
\begin{equation}
    \text{balance}(C_l) = \min_{s, s' \in [h], s \neq s'} \frac{|V_s \cap C_l|}{|V_{s'} \cap C_l|},
\end{equation}
where $|V_s \cap C_l|$ is the number of the members of group $V_s$ that also appear in cluster $C_l$. This metric measures the degree of the discrepancy in the proportional representation of each group $V_s$ in cluster $C_l$ by taking the minimum ratio across all pairs of groups. In other words, the balance of $C_l$ is determined by the largest difference in the proportional representation of each group $V_s$ in cluster $C_l$, i.e., the largest difference between $|V_s \cap C_l|$ and $|V_{s'} \cap C_l|$ among all pairs of groups. A cluster achieves perfect balance (value 1) only if all groups are equally represented in the cluster, while a lower balance value indicates that at least one group pair is unbalanced. The average balance is then obtained by taking the average of the balance values over all $k$ clusters.

\noindent \textbf{Hardware and software configuration.} For hardware, all experiments are conducted on a Linux server equipped with an Intel Xeon Gold 5218R CPU, 95GB RAM, and an NVIDIA Quadro RTX 8000 48GB GPU. For software, we use Python 3.12, SciPy v1.11, CuPy v13.2, and CUDA 12.3. Each experiment is repeated ten times, and the average results are reported for performance evaluation.

\subsection{Experiment Results}
\noindent \textbf{Synthetic dataset.} We first consider the mSBM dataset to demonstrate the effectiveness and scalability of FairAD in comparison with the baseline algorithms in terms of error rate and running time. As illustrative results, in Figure~\ref{fig:sbm1}, we present the results for the cases with $h = 2$ and $k= 4$, $h = 5$ and $k = 5$, and $h = 10$ and $k = 5$. Here, the graph sizes vary from $n = 5 \times 10^3$ to $n = 3\times 10^4$. As shown in Figure~\ref{fig:sbm1}, we observe that FairSC, sFairSC, and FairAD accurately obtain the ground truth clustering labels, whereas SC fails with a high error rate. While FairSC, sFairSC, and FairAD do the job correctly, FairAD is significantly faster than FairSC and sFairSC in terms of running time, with the speed-ups of up to $42\times$ and $12\times$, respectively.

\begin{figure*}[!t]
    \setlength{\subfigcapskip}{-4pt}
  \centering

  \subfigure{%
    \includegraphics[width=0.4\textwidth]{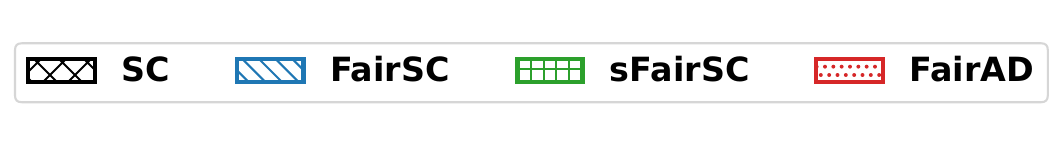}%
    \addtocounter{subfigure}{-1}
  }\\[-5mm]  

  \subfigure[NBA]{%
    \includegraphics[width=0.32\textwidth]{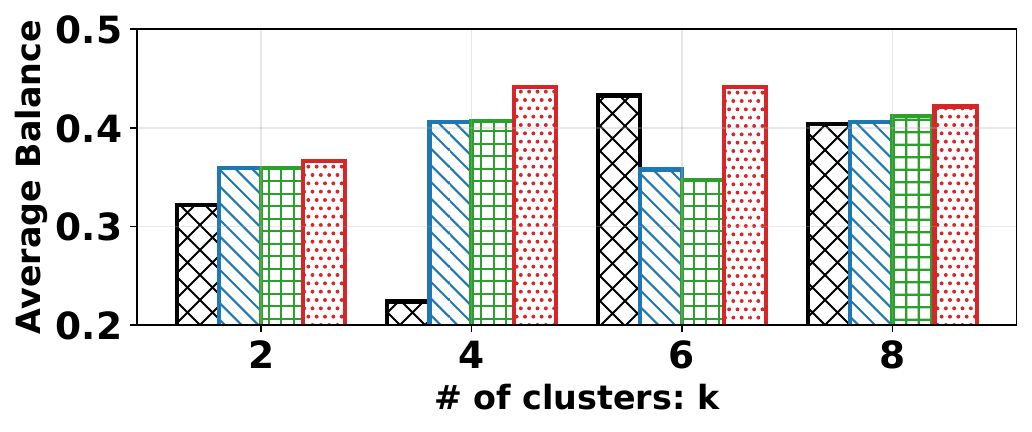}%
    \label{fig:nba_balance}
  }\hfill
  \subfigure[German]{%
    \includegraphics[width=0.32\textwidth]{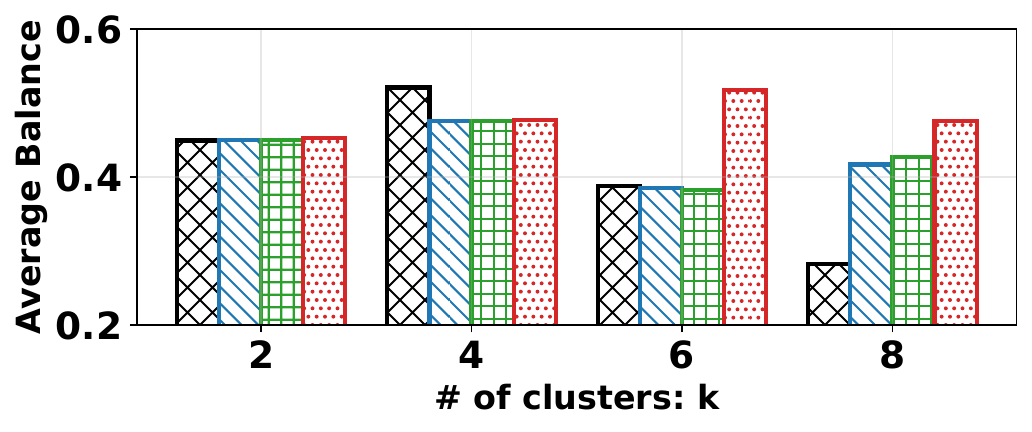}%
    \label{fig:german_balance}
  }\hfill
  \subfigure[LastFM]{%
    \includegraphics[width=0.32\textwidth]{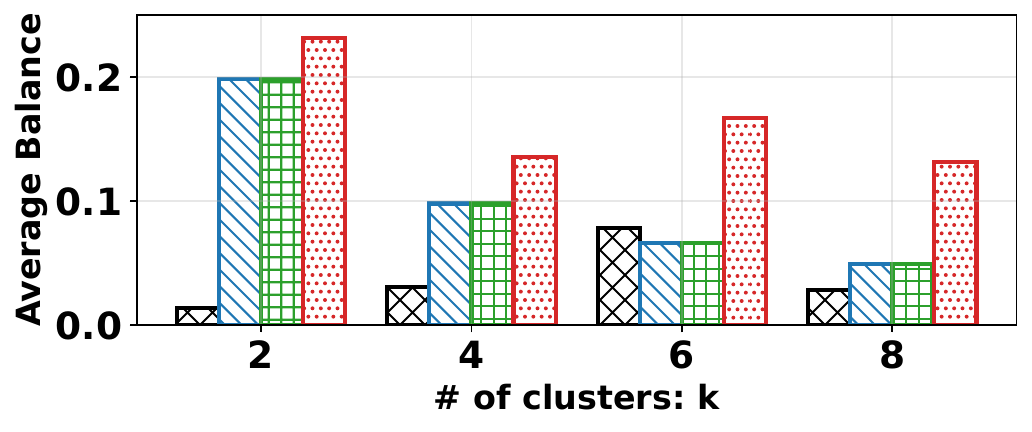}%
    \label{fig:lastfm_balance}
  }
  \\
  \vspace{-3mm}
    \subfigure[Recidivism]{%
    \includegraphics[width=0.32\textwidth]{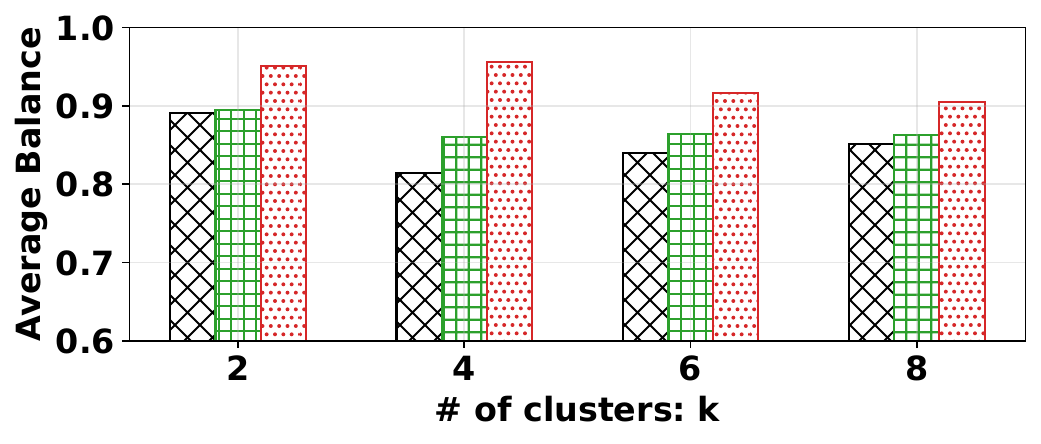}%
    \label{fig:Recividism_balance}
  }\hfill
  \subfigure[Deezer]{%
    \includegraphics[width=0.32\textwidth]{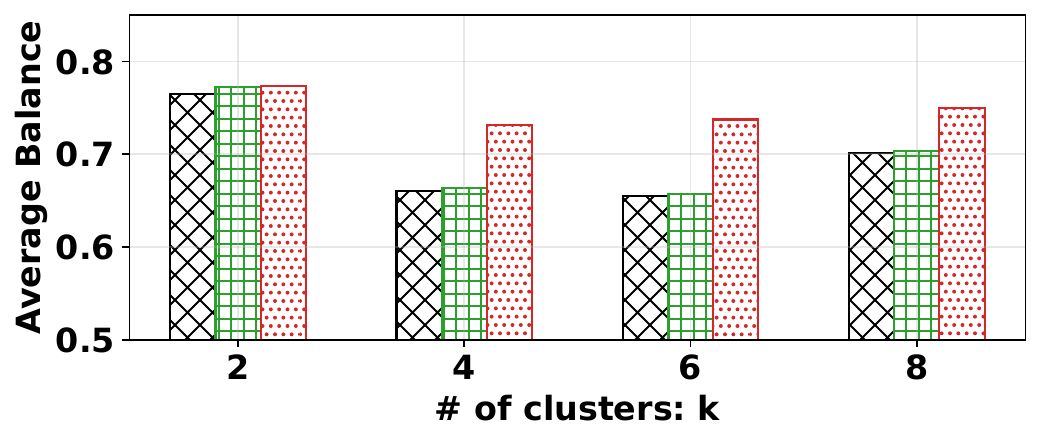}%
    \label{fig:deezer_balance}
  }\hfill
  \subfigure[Credit]{%
    \includegraphics[width=0.32\textwidth]{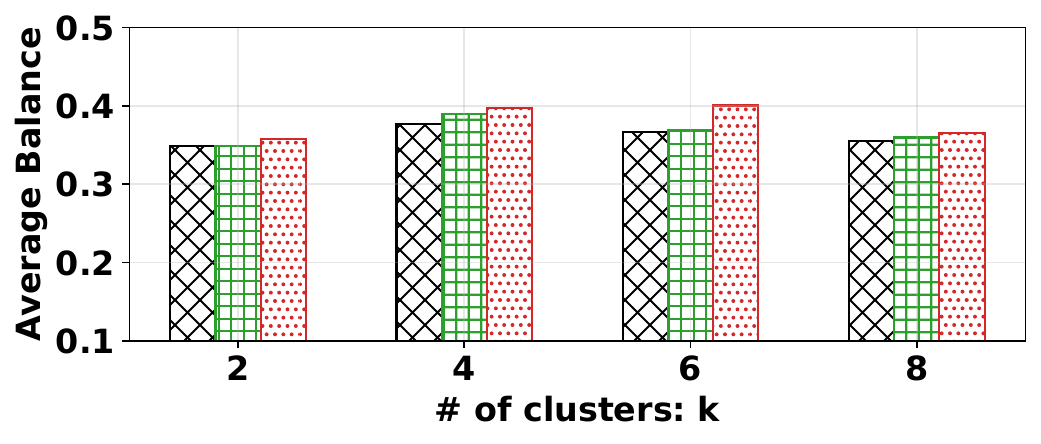}%
    \label{fig:credit_balance}
  }
    \vspace{-5mm}
  \caption{%
   Average balance for NBA, German, and LastFM datasets (top row) and for Recidivism, Deezer, and Credit datasets (bottom row), when changing the number of clusters.%
  }
  \label{fig:combined_balance}
  \vspace{-3mm}
\end{figure*}

\begin{figure*}[!t]
\setlength{\subfigcapskip}{-4pt}
  \centering

  \subfigure{%
    \includegraphics[width=0.4\textwidth]{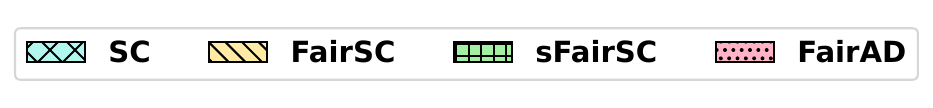}%
 \addtocounter{subfigure}{-1}
  }\\[-4mm]  

  \subfigure[NBA]{%
    \includegraphics[width=0.32\textwidth]{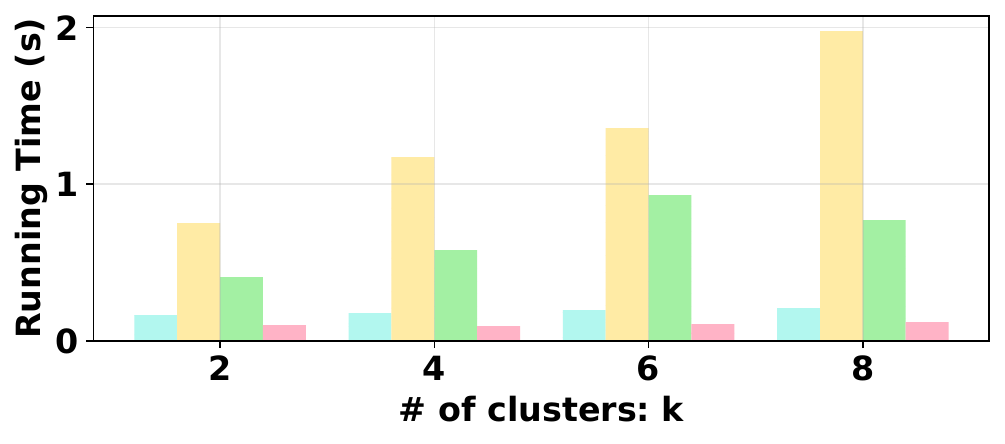}%
    \label{fig:nba_time}
  }\hfill
  \subfigure[German]{%
    \includegraphics[width=0.32\textwidth]{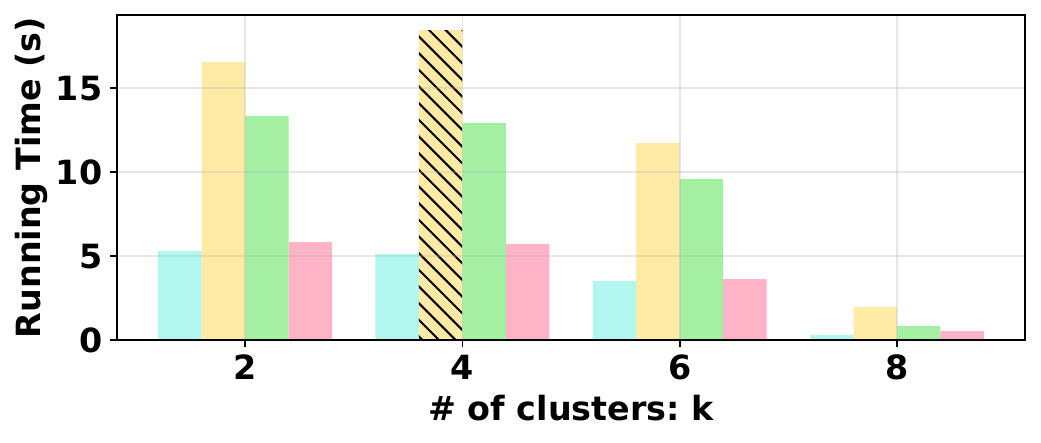}%
    \label{fig:german_time}
  }\hfill
  \subfigure[LastFM]{%
    \includegraphics[width=0.32\textwidth]{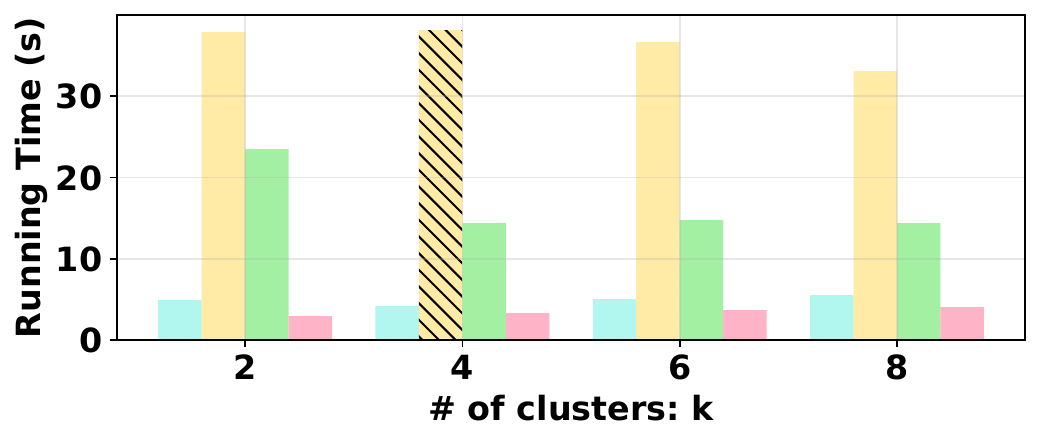}%
    \label{fig:lastfm_time}
  }
  \\
  \vspace{-3mm}
    \subfigure[Recidivism]{%
    \includegraphics[width=0.32\textwidth]{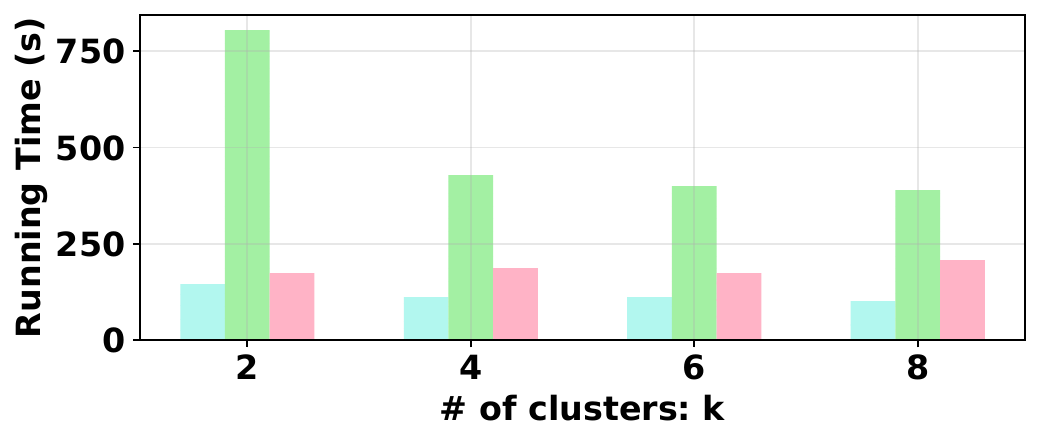}%
    \label{fig:Recividism_time}
  }\hfill
  \subfigure[Deezer]{%
    \includegraphics[width=0.32\textwidth]{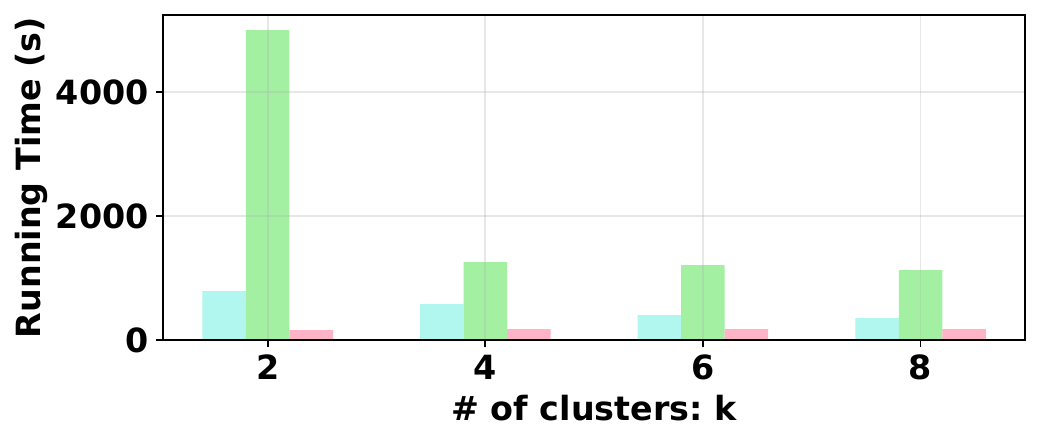}%
    \label{fig:Deezer_time}
  }\hfill
  \subfigure[Credit]{%
    \includegraphics[width=0.32\textwidth]{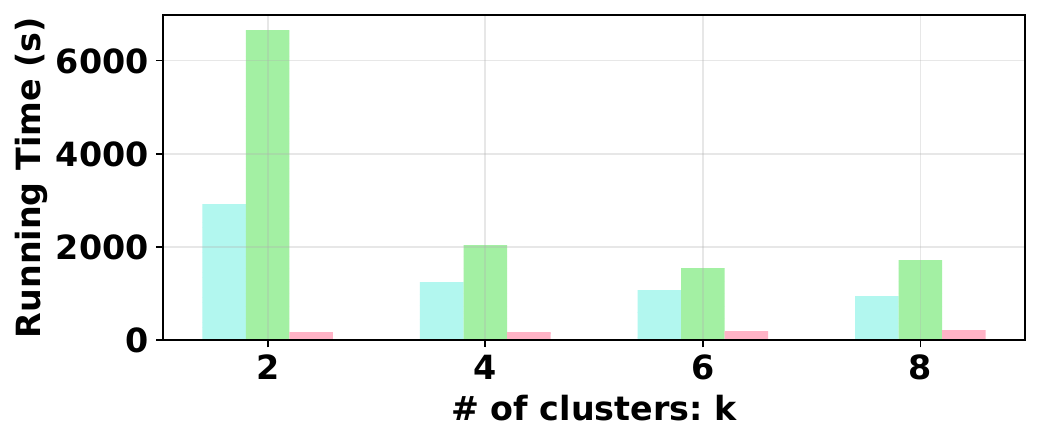}%
    \label{fig:Credit_time}
  }
    \vspace{-5mm}
  \caption{Running time for NBA, German, and LastFM datasets (top row) and for Recidivism, Deezer, and Credit datasets (bottom row), when changing the number of clusters.}
  \label{fig:combined_runtime}
  \vspace{-1mm}
\end{figure*}

We further evaluate the impact of varying values of $h$ and $k$ on the performance of FairAD and baselines when the graph sizes are $n = 20000$ and $n= 30000$. We observe that the error rates of FairSC, sFairSC, and FairAD are all zero for most cases, whereas SC's error rate ranges from 0.17 to 0.83. We omit the results for brevity. We here mainly evaluate the running time of each method. As shown in Tables \ref{tab:performance} and \ref{tab:performance2}, the running time of FairAD is just under 200 seconds, achieving over $10\times$ speed-up compared to sFairSC and $30\times$ to $40\times$ speed-up compared to FairSC for all test cases. While SC achieves the similar running time as FairAD, it has a much higher error rate. These results demonstrate the superior performance of FairAD across different settings.

\noindent \textbf{Real-world datasets.} We next evaluate the effectiveness and efficiency of FairAD on six real-world datasets in terms of the average balance and running time and demonstrate its superiority to the baselines. As shown in Figure~\ref{fig:combined_balance}, we observe that FairAD outperforms all baselines consistently in terms of average balance. Specifically, for the small graphs such as NBA, German, and LastFM, FairAD achieves an average improvement of approximately $20\%$, with up to about $100\%$ improvement, when compared to FairSC and sFairSC. Note that FairSC and sFairSC exhibit nearly identical balance performance for all graphs. We also observe that FairSC takes an excessive amount of time, i.e., several hours to days, for the larger graphs such as Recidivism, Deezer, and Credit, for which we do not report its results. For the larger graphs, FairAD outperforms sFairSC by $10\%$ to $15\%$. Overall, the results demonstrate the effectiveness of FairAD, which consistently achieves the highest average‑balance score for all test cases on all datasets. In other words, FairAD produces the most balanced clusters for all cases.

To evaluate the efficiency of FairAD,  we compare its running time with that of each baseline and report the results in Figure~\ref{fig:combined_runtime}. FairAD is at least twice as fast as sFairSC and more than $3\times$ faster than FairSC for all values of $k$ on the small graphs such as NBA, German, and LastFM. The performance gap in running time becomes wider on the larger graphs, as shown in Figures~\ref{fig:Recividism_time}--\ref{fig:Credit_time}. FairAD is more than an order of magnitude faster than sFairSC, achieving up to $40\times$ speed‑up compared to sFairSC. This is because sFairSC, especially its eigensolver, requires a much larger number of iterations to converge for such large graphs. We also observe that FairAD is even faster than SC for Deezer and Credit datasets, where SC already fails to produce balanced clusters.

To summarize, the experiments on both synthetic and real‑world datasets confirm the effectiveness and efficiency of FairAD.  It produces highly balanced clusters while taking only a fraction of the running times of its competing methods, namely FairSC and sFairSC, making it a practical solution for fair graph clustering.

\section{Conclusion}
We have introduced FairAD, a novel fair graph clustering method via algebraic distance. The main enabler of FairAD is its framework to impose fairness constraints into the affinity matrix when it is constructed based on the algebraic distance. FairAD then effectively leveraged graph coarsening to convert the optimization problem into a simpler graph cut problem, which is solved efficiently. Its implementation was further optimized through several techniques. Experiment results demonstrated the superior performance of FairAD to state-of-the-art fair graph clustering algorithms in terms of both the quality of fairness and computational efficiency.

\begin{acks}
This work was supported by the National Science Foundation under grants IIS-2209921, CNS-2209922, and DMS-2208499, the International Energy Joint R\&D Program of the Korea Institute of Energy Technology Evaluation and Planning (KETEP), granted financial resource from the Ministry of Trade, Industry \& Energy, Republic of Korea (No. 20228530050030), and an equipment donation from NVIDIA Corporation. C. Lee is the corresponding author.
\end{acks}

\section*{GenAI Usage Disclosure}

No GenAI tools were used in any stage of the research, nor in the writing.

\bibliographystyle{ACM-Reference-Format}
\balance
\bibliography{ref}

\end{document}